\documentclass[conference]{IEEEtran}
\IEEEoverridecommandlockouts
\usepackage{cite}
\usepackage{amsmath,amssymb,amsfonts}
\usepackage{algorithmic}
\usepackage{graphicx}
\usepackage{textcomp}
\usepackage{xcolor}

\usepackage{algorithm}
\usepackage{algorithmic}

\usepackage{amsmath}
\usepackage{amssymb}
\usepackage{mathtools}
\usepackage{amsthm}

\usepackage{multirow}   

\usepackage{caption}
\usepackage{subfig}

\def\BibTeX{{\rm B\kern-.05em{\sc i\kern-.025em b}\kern-.08em
    T\kern-.1667em\lower.7ex\hbox{E}\kern-.125emX}}
\begin{document}

\title{Latent Space Score-based Diffusion Model for Probabilistic Multivariate Time Series Imputation\\

}

\author{\IEEEauthorblockN{Guojun Liang}
\IEEEauthorblockA{\textit{School of Information Technology } \\
\textit{Halmstad University}\\
Halmstad, Sweden \\
guojun.liang@hh.se}
\and
\IEEEauthorblockN{Najmeh Abiri}
\IEEEauthorblockA{\textit{School of Information Technology} \\
\textit{Halmstad University}\\
Halmstad, Sweden \\
najmeh.abiri@hh.se}
\and
\IEEEauthorblockN{Atiye Sadat Hashemi}
\IEEEauthorblockA{\textit{School of Information Technology$^{1}$} \\
\textit{Halmstad University$^{1}$}\\
\textit{Lund University$^{2}$}\\
atiye-sadat.hashemi@hh.se}
\and
\IEEEauthorblockN{Jens Lundström}
\IEEEauthorblockA{\textit{School of Information Technology} \\
\textit{Halmstad University}\\
Halmstad, Sweden \\
jens.r.lundstrom@hh.se}
\and
\IEEEauthorblockN{Stefan Byttner}
\IEEEauthorblockA{\textit{School of Information Technology} \\
\textit{Halmstad University}\\
Halmstad, Sweden \\
stefan.byttner@hh.se}
\and
\IEEEauthorblockN{Prayag Tiwari}
\IEEEauthorblockA{\textit{School of Information Technology} \\
\textit{Halmstad University}\\
Halmstad, Sweden \\
prayag.tiwari@hh.se}
}

\maketitle

\begin{abstract}
Accurate imputation is essential for the reliability and success of downstream tasks. Recently, diffusion models have attracted great attention in this field. However, these models neglect the latent distribution in a lower-dimensional space derived from the observed data, which limits the generative capacity of the diffusion model. Additionally, dealing with the original missing data without labels becomes particularly problematic. To address these issues, we propose the Latent Space Score-Based Diffusion Model (LSSDM) for probabilistic multivariate time series imputation. Observed values are projected onto low-dimensional latent space and coarse values of the missing data are reconstructed without knowing their ground truth values by this unsupervised learning approach. Finally, the reconstructed values are fed into a conditional diffusion model to obtain the precise imputed values of the time series. In this way, LSSDM not only possesses the power to identify the latent distribution but also seamlessly integrates the diffusion model to obtain the high-fidelity imputed values and assess the uncertainty of the dataset. Experimental results demonstrate that LSSDM achieves superior imputation performance while also providing a better explanation and uncertainty analysis of the imputation mechanism. The website of the code is \textit{https://github.com/gorgen2020/LSSDM\_imputation}.
\end{abstract}


\section{Introduction}
Multivariate time series imputation (MTSI) is a crucial approach for addressing missing data, and scholars have developed a range of MTSI models from diverse perspectives and applications \cite{qiu2020multivariate} \cite{ding2024accurate}, including healthcare, transportation, etc. Leveraging the strengths of recurrent neural network (RNN) architectures, such as GRU and LSTM, Bi-directional GRU are employed in Brits \cite{cao2018brits} for MTSI tasks. Some researchers state that the time series owns the latent graph structure \cite{pakiyarajah2024irregularity} and utilizes a Graph Neural Network (GNN) to implement the imputation by its neighbor's features \cite{viswarupan2024mixed}. Grin \cite{cinifilling} is developed to deal with the graph structure by GNN with bidirectional GRU. HSPGNN \cite{liang2024physics} incorporates the physics law into the GNN structure and tries to obtain the missing values by the physics law. With the development of the transformer structure \cite{vaswani2017attention}, attention is widely used in some imputation models \cite{xu2022pulseimpute}. SAITS \cite{du2023saits} employs a self-supervised training scheme with two diagonal-masked self-attention blocks and a weighted-combination block to impute missing data. Imputation methods can be broadly categorized into two types: discriminative and generative approaches. Most discriminative models belong to supervised learning, which needs access to the ground truth values for training. However, the requirement can not be satisfied in most real scenarios since most datasets suffer from the original missing phenomenon. They over-rely on the ground truth of the missing values and are easily affected by the noise, thereby not accounting for the uncertainty in the imputed values. Generative Deep Learning Methods, based on their ability to yield varied imputations that reflect the inherent uncertainty in the imputation \cite{abiri2019establishing, choi2023conditional}, are research hotspots in MTSI. GP-VAE \cite{fortuin2020gp} employs the convolutional neural network (CNN) and VAE to find out the latent distribution by maximizing the Evidence Lower Bound (ELBO) on the marginal likelihood. Even though it is an efficient, interpretable latent space, it has smoother and potentially less detailed outputs. The diffusion model \cite{alcarazdiffusion} has attracted a lot of attention due to its high-quality output and better theoretical strength and robustness. CSDI \cite{tashiro2021csdi} imputes the missing values based on the conditional diffusion probability model, which treats different nodes as multiple features of the time series and uses a Transformer to capture feature dependencies. PriSTI \cite{liu2023pristi} imputes missing values with the help of the extracted conditional feature by a conditional diffusion framework to calculate temporal and spatial global correlations, but this method is hard to apply to the health care datasets such as electronic health records (EHRs) since there are no apparent spatial global correlation and geographic dependency of this irregular healthcare datasets. Nonetheless, these diffusion models do not consider the latent distribution of the dataset, which limits the further generative capability and the explanation of the imputation mechanism. Also, the imputation performance degrades dramatically when the original missing data has a significant proportion of the dataset since they do not have the ground truth of the original missing data to add noise to these models. To address the questions mentioned above, we propose the Latent Space Score-Based Diffusion Model (LSSDM) for probabilistic multivariate time series imputation to unify the unsupervised and supervised generative models of VAE and diffusion in MTSI. LSSDM not only possesses the power to identify the latent distribution but also seamlessly integrates the diffusion model to obtain the high-fidelity imputed values and assess the uncertainty of the dataset. Moreover, the unsupervised learning VAE framework can handle the original missing data and the simulated missing data effectively without knowing the ground truth values, which makes these models more feasible in real scenarios.

\section{Methodology}
\label{sec:format}
\subsection{Preliminary}
As for the MTSI problem, we denote $\mathbf{X}_0^{N \times D}$ as the input feature matrix, where $N$ is the number of sensors and $D$ is the length of the time series. For the imputation task, $\mathbf{M} \in \mathbb{R}^{N \times D} $ is used to indicate the missing mask, where $\mathbf{M}_{ij}=0$ means that the $j$-th sensor at time step $i$-th is missing, otherwise, $\mathbf{M}_{ij}=1$. For Facilitating subsequent discussions, $\mathbf{X}_0^{co} = \mathbf{X} \odot \mathbf{M}$ ($\odot$ is the Hadamard product) is adopted as the conditional observed features while $\mathbf{X}_0^{ta} = \mathbf{X} \odot (1 -\mathbf{M})$ as the missing target. Moreover, $\mathbf{A}$ represents the adjacent matrix if the time series own the graph structure, i.e., the traffic domain.

\subsection{Neural network architecture}
\begin{figure}[!t] 
 \centering 
\includegraphics[width=\columnwidth]{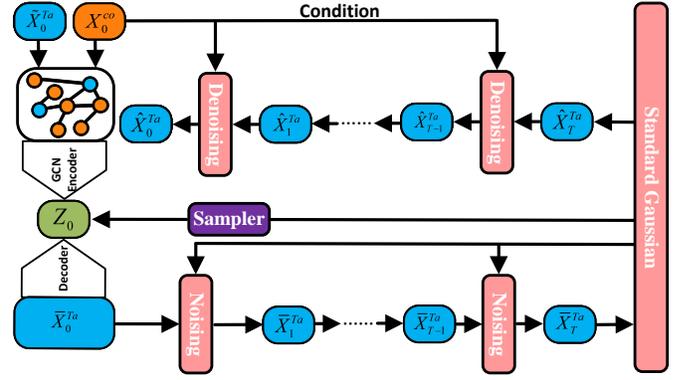}  
\caption{The framework of LSSDM.}
\label{framework}   
\end{figure}

The Framework of LSSDM is shown in Fig \ref{framework}. Firstly, the conditional observed values $\mathbf{X}_0^{co}$ together with the coarse values of the linear interpolation $\mathbf{\tilde{X}}^{ta}$ are projected onto the stochastic variable of low dimension latent space $\mathbf{Z}_0$ by the graph convolutional neural network (GCN) structure. Secondly, the reconstructed imputed values $\mathbf{\Bar{X}}_0^{ta}$ are calculated by the transformer structure in the decoder stage. Then, $\mathbf{\Bar{X}}_0^{ta}$ are fed to the forward diffusion process by adding the noise. Finally, we adopt DiffWave \cite{kong2020diffwave} to learn the noise in the denoising process, which is conditional on the observed values $\mathbf{X}_0^{co}$. Finally, the accurate imputed values $\mathbf{\Hat{X}}_0^{ta}$ can be obtained by sampling from the standard Gaussian noise through the learned denoising neural network.

To determine the precise imputation value and latent distribution in the projected space, we apply Bayes' rule to compute the probability distribution of the dataset, $\log p(\mathbf{X}_0)$ as follows:
\begin{equation}
    \begin{aligned}
        & \log p(\mathbf{X}_0)=\log p(\mathbf{X}_0^{co},\mathbf{X}_0^{ta})=\log p(\mathbf{X}_0^{ta}|\mathbf{X}_0^{co}) +\underbrace{\log p(\mathbf{X}_0^{co})}_{Evidence} \\
        & \approx  \log [\int p(\mathbf{X}_0^{ta}|\mathbf{X}_0^{co}, \mathbf{\Bar{X}}_0^{ta})p(\mathbf{X}_0^{co}, \mathbf{\Bar{X}}_0^{ta}|\mathbf{Z}_0)p(\mathbf{Z}_0) \mathrm{d} \mathbf{Z}_0] \\
        & = \log \int p(\mathbf{X}_0^{co}, \mathbf{\Bar{X}}_0^{ta}|\mathbf{Z}_0)p(\mathbf{Z}_0) \mathrm{d} \mathbf{Z}_0  +\log p(\mathbf{X}_0^{ta}|\mathbf{X}_0^{co}, \mathbf{\Bar{X}}_0^{ta}).\\
    \end{aligned} 
    \label{total_objective}
\end{equation}
The target function can thus be decomposed into two terms. For the first term, considering the graph structure, we can maximize the ELBO using the Variational Graph Autoencoder (VGAE) algorithm. The VGAE is trained to accurately reconstruct missing data from the latent representation without access to the ground truth of the missing values, as it is an unsupervised generative model. To mitigate the sparsity of the input features, during the training stage, we use linear interpolation to estimate coarse values $\mathbf{\tilde{X}}^{ta}$ for all missing values, including both the original and simulated missing values. The deduction is shown as follows:
\begin{equation}
\begin{aligned}
             &\log \int p(\mathbf{X}_0^{co}, \mathbf{\Bar{X}}_0^{ta}|\mathbf{Z}_0)p(\mathbf{Z}_0) \mathrm{d} \mathbf{Z}_0\ge 
          \mathbb{E}_{q_{\phi}(\mathbf{Z}_{0}|\mathbf{X}_0^{co}, \mathbf{\tilde{X}}_0^{ta}, \mathbf{A})}  \\
          & \log p_{\psi}(\mathbf{X}_0^{co}, \mathbf{\Bar{X}}_0^{ta}|\mathbf{Z}_{0})
             -\textbf{KL}[q_{\phi}(\mathbf{Z}_{0}|\mathbf{X}_0^{co}, \mathbf{\tilde{X}}_0^{ta}, \mathbf{A})||p(\mathbf{Z}_{0})],\\
\end{aligned}
\end{equation}
where $\phi$ and $\psi$ are the learnable parameters. we assume the posterior distribution obey the Gaussian distribution $q_{_{\phi}}(\mathbf{Z}_{0}|\mathbf{X}_0^{co}, \mathbf{\tilde{X}}_0^{ta}, \mathbf{A}) = \mathcal{N}(\mathbf{u}_{\phi}, \mathbf{\Sigma}_{\phi})$. Then, the posterior distribution can be estimated by GCN model \cite{kipf2016semi}:
\begin{equation}
\label{eq4} 
\mathbf{H}^{(l+1)}=GCN\left(\mathbf{A}, \mathbf{H}^{(l)}\right)=\sigma\left( \mathbf{L} \mathbf{H}^{(l)} \mathbf{w}^{(l)}\right),
\end{equation}
where the Laplacian matrix can be formulated as $\mathbf{L}=\mathbf{D}^{-\frac{1}{2}} (\mathbf{I}-\mathbf{A})\mathbf{D}^{\frac{1}{2}}$, where $\mathbf{I}$ is the identity matrix, which means adding self-connections, $\mathbf{D}$ is the degree matrix. where $\mathbf{H}^{(l)} \in \mathbb{R}^{N \times F}$ is the input of $l$ layer, while $\mathbf{H}^{(l+1)} \in \mathbb{R}^{N \times E}$ is output of $l$ layer with $E$ embedding. In addition, $\mathbf{w}^{(l)} \in \mathbb{R}^{F \times E}$ is a learnable parameter, and $\sigma$ represents the nonlinear activation function. In this study, we adopt the 2-layer GCN as the encoder framework. The mean and variance share the first-layer parameters but are independent in the second layer. Thus, the output of the latent variable can be expressed as:
\begin{equation}
\scalebox{0.85}{$
\begin{aligned}
        &\mathbf{u}_\phi = GCN(\mathbf{X},\mathbf{A})=Sigmoid (\mathbf{L} (ReLU(\mathbf{LXW^{(1)}))W^{(2)}_u})\\
        &\log \mathbf{\Sigma}_\phi = GCN(\mathbf{X},\mathbf{A}) = Sigmoid (\mathbf{L} (ReLU(\mathbf{LXW^{(1)}))W^{(2)}_{\Sigma}}).\\
\end{aligned}$}
\label{GCN}
\end{equation}

The latent space distribution is obtained through the GCN. During the reconstruction stage, we utilize a combination of a Transformer and 1D CNN to reconstruct the observed data, represented as $f_{\psi}$ as follows:
\begin{equation}
    \mathbf{\Bar{X}}_0 = f_{\psi} (\mathbf{Z}_0)= \textit{CNN}(\textit{CNN}(\textit{transformer}(\mathbf{Z}_0))).
    \label{reconstructed_function}
\end{equation}
where $\psi$ represents the learnable parameters of the CNN and Transformer. Assuming that the prior latent distribution follows a standard Gaussian distribution, $p(\mathbf{Z}_0) =  \mathcal{N}(0,\mathbf{I})$, where $\mathbf{I}$ is the Identity matrix and ignoring the constant term, the objective function can be reformulated using the reparameterization trick as follows:
\begin{equation}
    \begin{aligned}
        \mathcal{L}_1  =&\mathbb{E}_{\epsilon \sim \mathcal{N}(0,\mathbf{I})} ||\mathbf{X}_0^{co} - f_\psi(\mathbf{u}_{\phi} + \mathbf{\Sigma}_{\phi }* \epsilon)||_{2}\odot \mathbf{M}\\
        &-\frac{1}{2} (tr(\mathbf{\Sigma}_{\phi})+\mathbf{u}_{\phi}^T\mathbf{u}_{\phi} \log det (\mathbf{\Sigma}_{\phi})),\\
    \end{aligned}
    \label{vae_loss}
\end{equation}
Where $\epsilon$ is the standard Gaussian noise and $tr$ is the matrix trace while $det$ is the determinant of the matrix. Consequently, the reconstructed values $\mathbf{\Bar{X}}^{ta}$, generated by the VGAE model, will serve as input for the diffusion model in the second term of Eq. \ref{total_objective}. The reconstructed values $\mathbf{\Bar{X}}^{ta}$ are used to calculate the probability of $p(\mathbf{X}_0^{ta}|\mathbf{X}_0^{co}, \mathbf{\Bar{X}}_0^{ta})$ within the diffusion framework. During the diffusion process, noise is added to $\mathbf{\Bar{X}}_0^{ta}$ for $T$ time step, which can be formulated as:
\begin{equation}
\begin{aligned}
         & q(\mathbf{\Bar{X}}_{1:T}^{ta}|\mathbf{X}_0^{co}, \mathbf{\Bar{X}}_0^{ta})=\prod_{t=1}^T  q(\mathbf{\Bar{X}}_{t}^{ta}|\mathbf{\Bar{X}}_{t-1}^{ta}, \mathbf{X}_0^{co}, \mathbf{\Bar{X}}_0^{ta})\\
        \textit{and} \quad & q(\mathbf{\Bar{X}}_{t}^{ta}|\mathbf{\Bar{X}}_{t-1}^{ta}, \mathbf{X}_0^{co}, \mathbf{\Bar{X}}_0^{ta})=\mathcal{N}(\sqrt{\alpha_t} \mathbf{\Bar{X}}_{t-1}^{ta},(1-\alpha_t) \mathbf{I}),
\end{aligned}
\label{diffusion_forward}
\end{equation}
where $\alpha _t$ evolves over time according to a fixed value. In the denoising stage, the neural network $p_{\theta}$ is adopted to learn the noise and restore the signal from the standard Gaussian noise. Thus, as for the second term of Eq. \ref{total_objective}, we can also use the ELBO again as follows:
\begin{equation}
    \begin{aligned}
        & \scalebox{0.98}{$\log p(\mathbf{X}_0^{ta}|\mathbf{X}_0^{co}, \mathbf{\Bar{X}}_0^{ta}) \ge \mathbb{E}_{q(\mathbf{X}_{1:T}^{ta}|\mathbf{X}_0^{co}, \mathbf{\Bar{X}}_0^{ta})} \log \frac{p_\theta(\mathbf{X}_{0:T}^{ta}|\mathbf{X}_0^{co}, \mathbf{\Bar{X}}_0^{ta})}{q(\mathbf{X}_{1:T}^{ta}|\mathbf{X}_0^{co}, \mathbf{\Bar{X}}_0^{ta})} $}\\
        &= \mathbb{E}_{q} \log \frac{p_\theta(\mathbf{X}_{T}^{ta}|\mathbf{X}_0^{co}, \mathbf{\Bar{X}}_0^{ta})\prod_{t=1}^{T}p_\theta(\mathbf{X}_{t-1}^{ta}|\mathbf{X}_{t}^{ta}, \mathbf{X}_0^{co}, \mathbf{\Bar{X}}_0^{ta})}{\prod_{t=1}^{T}q(\mathbf{X}_{t-1}^{ta}|\mathbf{X}_{t}^{ta}, \mathbf{X}_0^{co}, \mathbf{\Bar{X}}_0^{ta})}\\
        & =\mathbb{E}_{q} \log \frac{p_\theta(\mathbf{X}_{T}^{ta}|\mathbf{X}_0^{co}, \mathbf{\Bar{X}}_0^{ta})p_\theta(\mathbf{X}_{0}^{ta}|\mathbf{X}_{1}^{ta}, \mathbf{X}_0^{co}, \mathbf{\Bar{X}}_0^{ta})
        }{q(\mathbf{X}_{T}^{ta}|\mathbf{X}_{t}^{ta}, \mathbf{X}_0^{co}, \mathbf{\Bar{X}}_0^{ta})} \\
        &+ \mathbb{E}_{q} \sum_{t=2}^{T}\log \frac{p_\theta(\mathbf{X}_{t-1}^{ta}|\mathbf{X}_{t}^{ta}, \mathbf{X}_0^{co}, \mathbf{\Bar{X}}_0^{ta})}{q(\mathbf{X}_{t-1}^{ta}|\mathbf{X}_{t}^{ta}, \mathbf{X}_0^{co}, \mathbf{\Bar{X}}_0^{ta})}.\\
    \end{aligned}
\end{equation}
Our goal is to learn the conditional distribution $p_\theta$ in the reverse process. As described in DDPM \cite{ho2020denoising}, the reverse process involves denoising $\mathbf{X}_{t}^{ta}$ to recover 
$\mathbf{X}_{0}^{ta}$, which is defined as:
\begin{equation}
\begin{aligned}
    & p_\theta(\mathbf{X}_{t-1}^{ta}|\mathbf{X}_{t}^{ta}, \mathbf{X}_0^{co}, \mathbf{\Bar{X}}_0^{ta}) \\
   & = \mathcal{N}(\mathbf{u}_\theta(\mathbf{X}_{t}^{ta}|\mathbf{X}_0^{co}, \mathbf{\Bar{X}}_0^{ta},t),\mathbf{\sigma}(\mathbf{X}_{t}^{ta}|\mathbf{X}_0^{co}, \mathbf{\Bar{X}}_0^{ta}, t) , \\
   & \scalebox{0.9}{$
   \mathbf{u}_\theta(\mathbf{X}_{t}^{ta}|\mathbf{X}_0^{co}, \mathbf{\Bar{X}}_0^{ta},t)=\frac{1}{\sqrt{\alpha_t}} \mathbf{X}_{t}^{ta}- \frac{1-\alpha}{\sqrt{1- \Bar{\alpha}_t}\sqrt{\alpha_t}}\epsilon_\theta(\mathbf{X}_{t}^{ta}|\mathbf{X}_0^{co}, \mathbf{\Bar{X}}_0^{ta},t), $}
\end{aligned}
\label{diffusion_u}
\end{equation}
where $\Bar{\alpha}_t = \prod_{i=1}^t \alpha_i$. From Eq. \ref{diffusion_forward} and \ref{diffusion_u}, with the parameterization, the second term of the objective function in Eq. \ref{total_objective} can be simplified to:
\begin{equation}
    \begin{aligned}
        &  \mathcal{L}_2 = \mathbb{E}_{\epsilon \sim \mathcal{N}(0,\mathbf{I})} (1- \mathbf{M}) \odot|| \epsilon -\epsilon_{\theta}(\mathbf{\Bar{X}}_{t}^{ta}|\mathbf{X}_0^{co}, \mathbf{\Bar{X}}_0^{ta},t))||_2 . \\
    \end{aligned}
    \label{diff_loss}
\end{equation}

To represent the reverse process, we use the same neural network framework as CSDI \cite{tashiro2021csdi}. Through the deduction mentioned above, the objective function of Eq. \ref{total_objective} can be transformed as:
\begin{equation}
\vspace{-0.1cm}
    \begin{aligned}
        &\mathcal{L}´=\mathcal{L}_1 +\mathcal{L}_2 = \mathbb{E}_{\epsilon \sim \mathcal{N}(0, \mathbf{I})} \left[ \left\| \mathbf{X}_0^{co} - f_\psi(\mathbf{u}_{\phi} + \mathbf{\Sigma}_{\phi} \cdot \epsilon) \right\|_{2} \odot \mathbf{M} \right. \\
    &\quad + \left. \left\| \epsilon - \epsilon_{\theta}(\mathbf{\Bar{X}}_{t}^{ta} \mid \mathbf{X}_0^{co}, \mathbf{\Bar{X}}_0^{ta}, t) \right\|_{2} \odot (1 - \mathbf{M}) \right]\\
    &-\frac{1}{2} (tr(\mathbf{\Sigma}_{\phi})+\mathbf{u}_{\phi}^T\mathbf{u}_{\phi} \log det (\mathbf{\Sigma}_{\phi}))\\
    \end{aligned}
\end{equation}
The details of the LSSDM algorithm are shown in Algorithm \ref{algorithm_LSSDM}. After training, we obtain the parameter $\mathbf{\lambda} =\{\phi, \psi, \theta \}$. Through the learned parameters, we can calculate more accurate missing values by sampling from the noise, and details of the total imputation algorithm are shown in Algorithm \ref{imputation}.

\begin{algorithm}
\caption{Training of LSSDM}
\begin{algorithmic}[hbt]
\STATE {\bfseries Input:} Time series values $\mathbf{X}_0^{co}$, mask $\mathbf{M}$ and Adjacent matrix $\mathbf{A}$ 
\STATE {\bfseries Output:} The model parameters $\mathbf{\lambda} =\{\phi, \psi, \theta \}$
\STATE Calculate the linear interpolation values $\mathbf{\tilde{X}}_0^{ta}$ and Laplacian matrix $\mathbf{L}$;
\STATE Initialize variables $\mathbf{\lambda}$;
\FOR{each epoch}
    \STATE Calculate the mean $\mathbf{u}_{\phi}$ and $\mathbf{\Sigma}_\phi$ variance by Eq. \ref{GCN};
    \STATE Optimize $\mathcal{L}_1$ in Eq, \ref{vae_loss} by gradient descent step;
    \STATE Obtain the reconstructed values $\mathbf{\Bar{X}}_{0}$ by Eq. \ref{reconstructed_function};
    \STATE Add noise to $\mathbf{\Bar{X}}_{0}^{ta}$ by Eq. \ref{diffusion_forward};
    \STATE Optimize $\mathcal{L}_2$ in Eq. \ref{diff_loss} by taking gradient step on\\
    $\mathbf{\nabla }_\theta  (1- \mathbf{M}) \odot|| \epsilon -\epsilon_{\theta}(\mathbf{\Bar{X}}_{t}^{ta}|\mathbf{X}_0^{co}, \mathbf{\Bar{X}}_0^{ta},t))||_2$;
\ENDFOR
\RETURN $\mathbf{\lambda}$ 
\end{algorithmic}
\label{algorithm_LSSDM}
\end{algorithm}

\vspace{-0.5cm}
\begin{algorithm}
\caption{Imputation of LSSDM}
\begin{algorithmic}[hbt]
\STATE {\bfseries Input:} Observed values $\mathbf{X}_0^{co}$, mask $\mathbf{M}$, Adjacent matrix $\mathbf{A}$ and $\mathbf{\lambda}= =\{\phi, \psi, \theta \}$
\STATE {\bfseries Output:} The predicted missing values $\mathbf{\hat{X}}_0^{ta}$
\STATE Calculate the linear interpolation values $\mathbf{\tilde{X}}_0^{ta}$;
\STATE Input $\mathbf{X}_0^{co}$ and $\mathbf{\tilde{X}}_0^{ta}$ to Eq. \ref{GCN} to obtain $\mathbf{u}_\phi$ and $\mathbf{\Sigma}_\phi$;
\STATE Sample noise $\epsilon$ and reparameterize by $\mathbf{Z}_0 = \mathbf{u}_{\phi} + \mathbf{\Sigma}_{\phi }* \epsilon$;
\STATE Obtain $\mathbf{\Bar{X}}_0^{ta}= (1-\mathbf{M})\odot \mathbf{\Bar{X}}_0$  by Eq. \ref{reconstructed_function};
\FOR{$t=T$ to 1}
\STATE Calculate $\mathbf{\Bar{X}}_t^{ta}$ by Eq. \ref{diffusion_forward};
\STATE Sample $\mathbf{\hat{X}}_{t-1}^{ta}$ using Eq. \ref{diffusion_u};
\ENDFOR
\RETURN  $\mathbf{\hat{X}}_0^{ta}$
\end{algorithmic}
\label{imputation}
\end{algorithm}

\section{Experiments and results}
To testify the performance of our model, PhysioNet 2012 \cite{silva2012predicting}, AQI-36 \cite{yi2016st}, and PeMS-BAY \cite{li2018dcrnn} datasets are used. AQI-36 is a benchmark for the time series imputation from the Urban Computing project in China. $PM2.5$ pollutant is collected hourly from May 2014 to April 2015 with various missing patterns and with 24.6\% originally missing rate. Four months (March, June, September, and December) are used as the test set without overlapping with the training dataset. PhysioNet 2012 (P12) is a healthcare dataset from the ICU that is 48 hours long with 4000 patients and 35 variables. PeMS-BAY is a famous traffic flow time series dataset, which was collected with 325 road sensors in the San Francisco Bay Area for nearly six months and re-sampled into 5 minutes by previous work \cite{li2018dcrnn}. For the P12 and PeMS-BAY datasets, 70\% of the total time series is used for training, while 10\% for validation and 20\% for evaluation. For AQI-36, we used the same mask as \cite{cinifilling}. As for the PeMS-BAY dataset, we adopt block missing and points missing masks as \cite{cinifilling} while  P12 only point missing as \cite{tashiro2021csdi}.
\subsection{Experimental setting}
The epoch and batch size are 200 and 16, respectively. Residual connection is applied, and the learning rate is 0.0001. The attention heads of the transformer adopt 8. For the diffusion stage, we adopt the same hyperparameters setting as CSDI \cite{tashiro2021csdi}. As for the P12 dataset, we use the Identity matrix as the Laplacian matrix since there is no pre-defined graph in this dataset. Notably, to highly reproduce real application scenarios, the simulated missing values are not allowed to take part in the training stage and are only used as the final evaluation, which is the same training and evaluation protocol as \cite{cinifilling, nie2023imputeformer} and different from \cite{tashiro2021csdi, liu2023pristi} since they allow it. Also, we only consider the out-of-sample scenarios, which means the training and evaluation series are disjoint sequences.

\subsection{Result}
The results of different datasets with different missing patterns are shown in Tab. \ref{maeresult}. The result indicates our model achieves the best imputation performance in the different missing situations under different datasets of different domains, which demonstrates the generalization and performance of our model. Interestingly, reconstructing the missing values from the projective low-dimension latent space can greatly improve the generative capability of the diffusion model, especially in dealing with the originally missing data effectively by this unsupervised learning framework without the demand of the ground truth values.
\vspace{-2.0em}
\begin{figure}[hbt] 
\subfloat[PeMS-BAY.]{ 
\includegraphics[width=0.48\columnwidth]{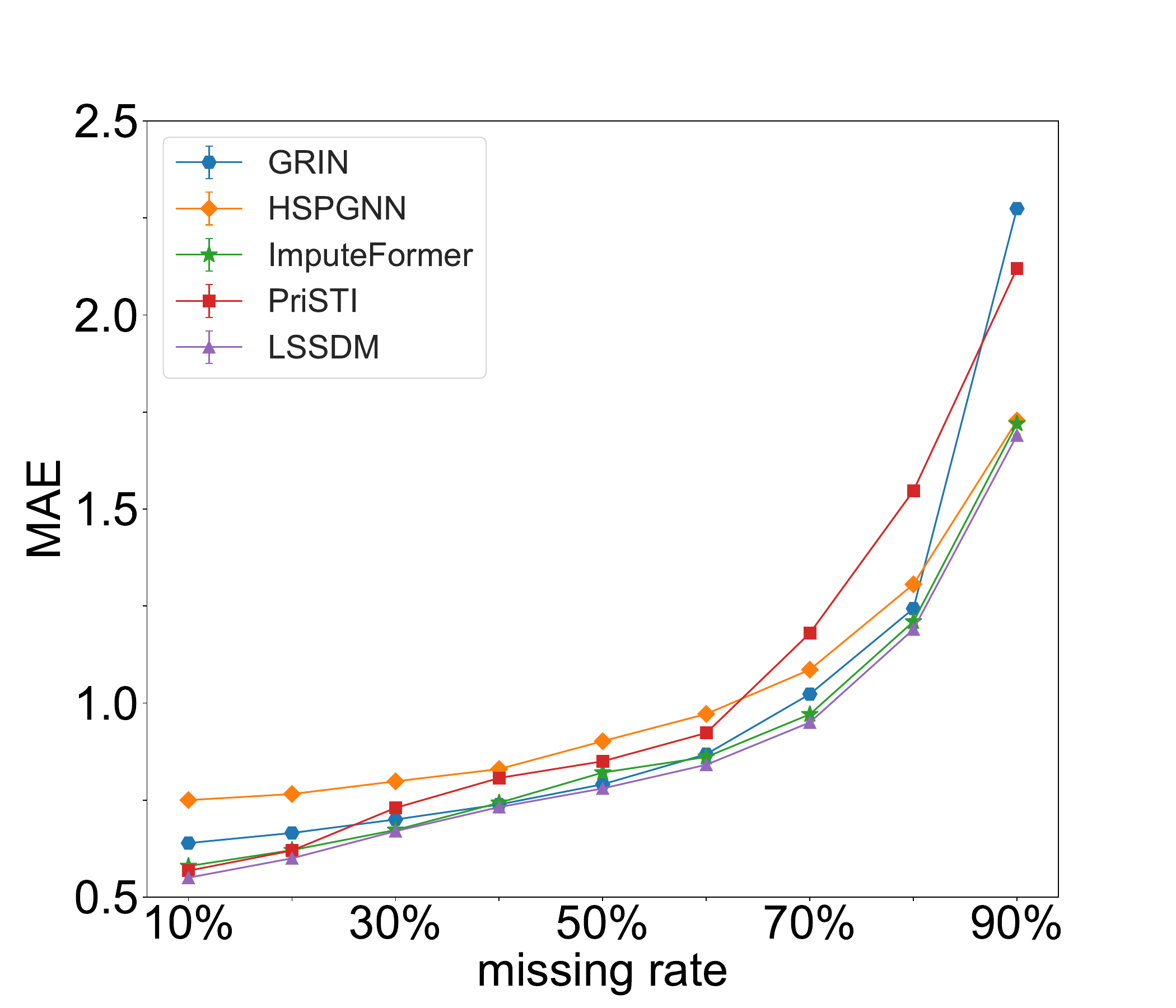} 
\label{MAE1} 
} 
\subfloat[P12.]{ 
\includegraphics[width=0.49\columnwidth]{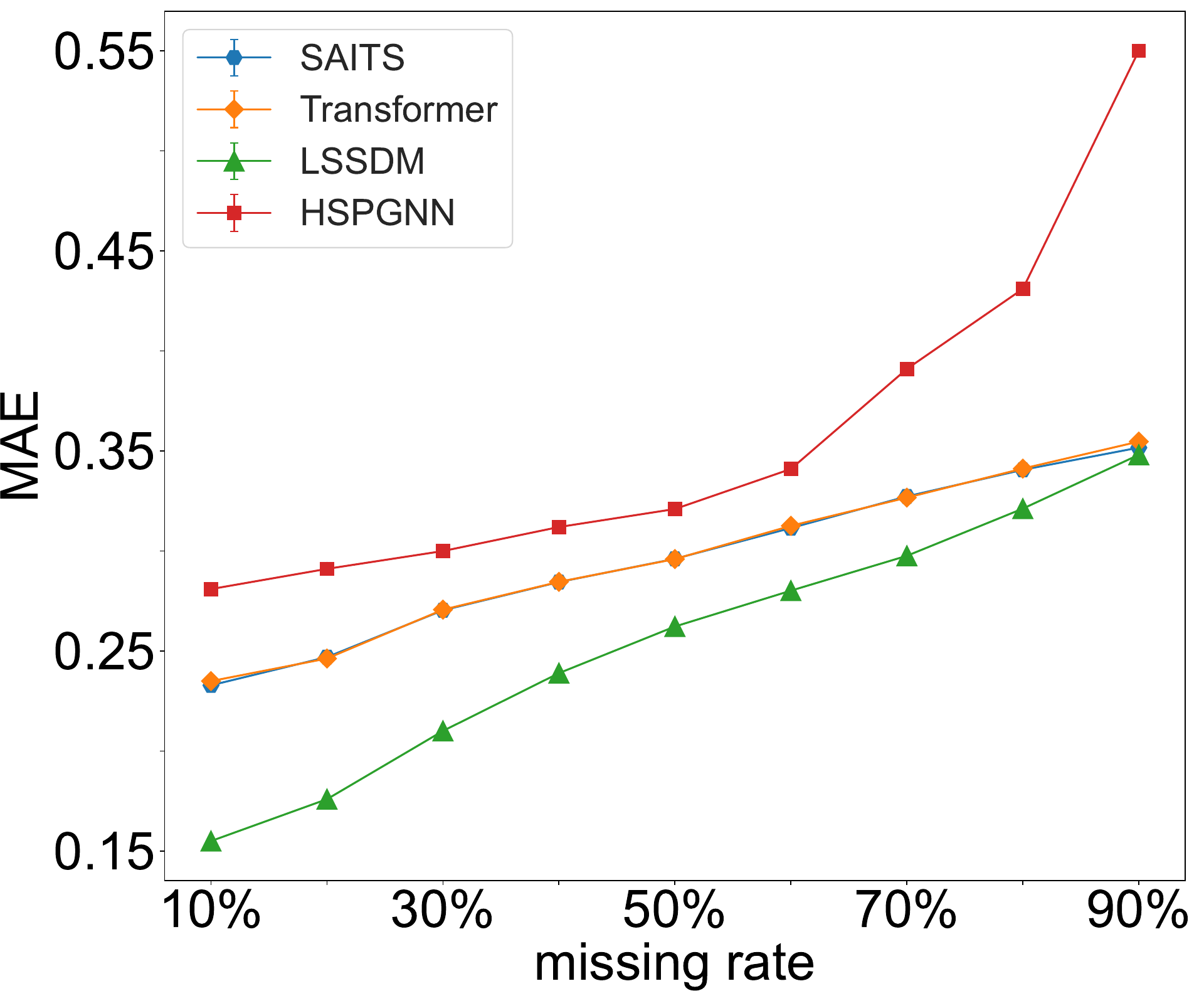}  
\label{MAE2} 
} 
\caption{MAE at different missing rates on different datasets.}
\label{missing_rate}
\end{figure} 
To investigate the performance of the SOTA models, we plot the MAE of PeMS-BAY and P12 under different missing rates. The results are shown in Fig. \ref{missing_rate}, which indicates that our model obtains the best robustness.
\begin{table}[hbt]
\centering
\caption{The imputation result of baselines.}
\scalebox{0.75}{
\renewcommand{\arraystretch}{1} 
\begin{tabular}{l ll ll ll}
\hline
\multirow{2}{*}{Model} & \multicolumn{2}{c}{AQI-36}                      & \multicolumn{2}{c}{\begin{tabular}[c]{@{}c@{}}P12\\ @50\%\end{tabular}} & \multicolumn{2}{c}{\begin{tabular}[c]{@{}c@{}}PeMS-BAY\\ @Block missing\end{tabular}} \\ \cline{2-7} 
                       & \multicolumn{1}{c}{MAE}        & \multicolumn{1}{c}{CPRS}           & \multicolumn{1}{c}{MAE}                      & \multicolumn{1}{c}{CPRS}                       & \multicolumn{1}{c}{MAE}                           & \multicolumn{1}{c}{CPRS}                             \\ 
Transformer \cite{vaswani2017attention}             & \multicolumn{1}{c}{12.00±0.60} & \multicolumn{1}{c}{---} & \multicolumn{1}{c}{0.297±0.002}                         &  \multicolumn{1}{c}{---}                     & \multicolumn{1}{c}{1.70±0.02 }                                &\multicolumn{1}{c}{---}                               \\ 

GRIN \cite{cinifilling}                   & \multicolumn{1}{c}{12.08±0.47} & \multicolumn{1}{c}{---}  & \multicolumn{1}{c}{0.403±0.003}                         & \multicolumn{1}{c}{---}                       & \multicolumn{1}{c}{1.14±0.01}                                &  \multicolumn{1}{c}{---}                             \\ 
SAITS \cite{du2023saits}                   & \multicolumn{1}{c}{18.13±0.35} & \multicolumn{1}{c}{---} & \multicolumn{1}{c}{0.296±0.002}              & \multicolumn{1}{c}{---}             & \multicolumn{1}{c}{1.56±0.01}                                & \multicolumn{1}{c}{---}                                \\ 
M$2$DMTF \cite{fan2021multi}                    & \multicolumn{1}{c}{14.61±0.49} & \multicolumn{1}{c}{---}   & \multicolumn{1}{c}{0.7002±0.001}                         & \multicolumn{1}{c}{---}                        & \multicolumn{1}{c}{2.49±0.02}                                &  \multicolumn{1}{c}{---}                                \\ 
HSPGNN  \cite{liang2024physics}                  & \multicolumn{1}{c}{11.19±0.20} & \multicolumn{1}{c}{---}    & \multicolumn{1}{c}{ 0.321±0.003  }                         &  \multicolumn{1}{c}{---}                           & \multicolumn{1}{c}{1.10±0.02  }                                &  \multicolumn{1}{c}{---}                             \\ 
ImputeFormer \cite{nie2023imputeformer}            & \multicolumn{1}{c}{11.58±0.20} & \multicolumn{1}{c}{---}    & \multicolumn{1}{c}{0.480±0.150}                         &  \multicolumn{1}{c}{---}                         & \multicolumn{1}{c}{0.95±0.02}                                & \multicolumn{1}{c}{---}                              \\ \hline
GP-VAE \cite{fortuin2020gp}                    & \multicolumn{1}{c}{25.71±0.30}           & \multicolumn{1}{c}{ 0.3377  }               & \multicolumn{1}{c}{0.511±0.007}    &    \multicolumn{1}{c}{0.7981}               & \multicolumn{1}{c}{2.86±0.15}                                & \multicolumn{1}{c}{0.0436}                                \\ 
PriSTI \cite{liu2023pristi}                   & \multicolumn{1}{c}{9.35±0.22}  & \multicolumn{1}{c}{0.1267}  & \multicolumn{1}{c}{0.611±0.006}                         & \multicolumn{1}{c}{0.5514}                      & \multicolumn{1}{c}{0.97±0.02}                                &  \multicolumn{1}{c}{0.0127} 	                              \\ 
CSDI \cite{tashiro2021csdi}                     & \multicolumn{1}{c}{9.57±0.10} & \multicolumn{1}{c}{0.1053} 
    & \multicolumn{1}{c}{0.660±0.002}                         & \multicolumn{1}{c}{0.5608}                        & \multicolumn{1}{c}{1.16±0.01  }                                &  \multicolumn{1}{c}{0.0135}                              \\ 
LSSDM (ours)            & \multicolumn{1}{c}{\textbf{ 8.89±0.23}}   & \multicolumn{1}{c}{\textbf{0.0969}}   & \multicolumn{1}{c}{\textbf{0.262±0.002}}                         &\multicolumn{1}{c}{\textbf{0.2890}}                        & \multicolumn{1}{c}{\textbf{0.89±0.02}}                                & \multicolumn{1}{c}{\textbf{0.0095}}         \\ \hline
\end{tabular}}
\label{maeresult}
\end{table}
\subsection{Uncertainty and latent state analysis} 
Unlike traditional deterministic imputation, which provides a single predicted value, probabilistic forecasts provide a distribution or range of possible outcomes along with their associated probabilities.  CRPS \cite{matheson1976scoring} measures how well the predicted cumulative distribution function (CDF) matches the observed outcome. In this study, we adopt the same measure as CSDI \cite{tashiro2021csdi}. We generate 100 samples to approximate the probability distribution as in the previous section. The result is shown in Tab. \ref{maeresult}. LSSDM outperforms the probabilistic baseline models, which better quantify the uncertainty accurately.

\begin{figure}[!hbt] \vspace{-1.2em}
\centering
\includegraphics[width=0.68\columnwidth]{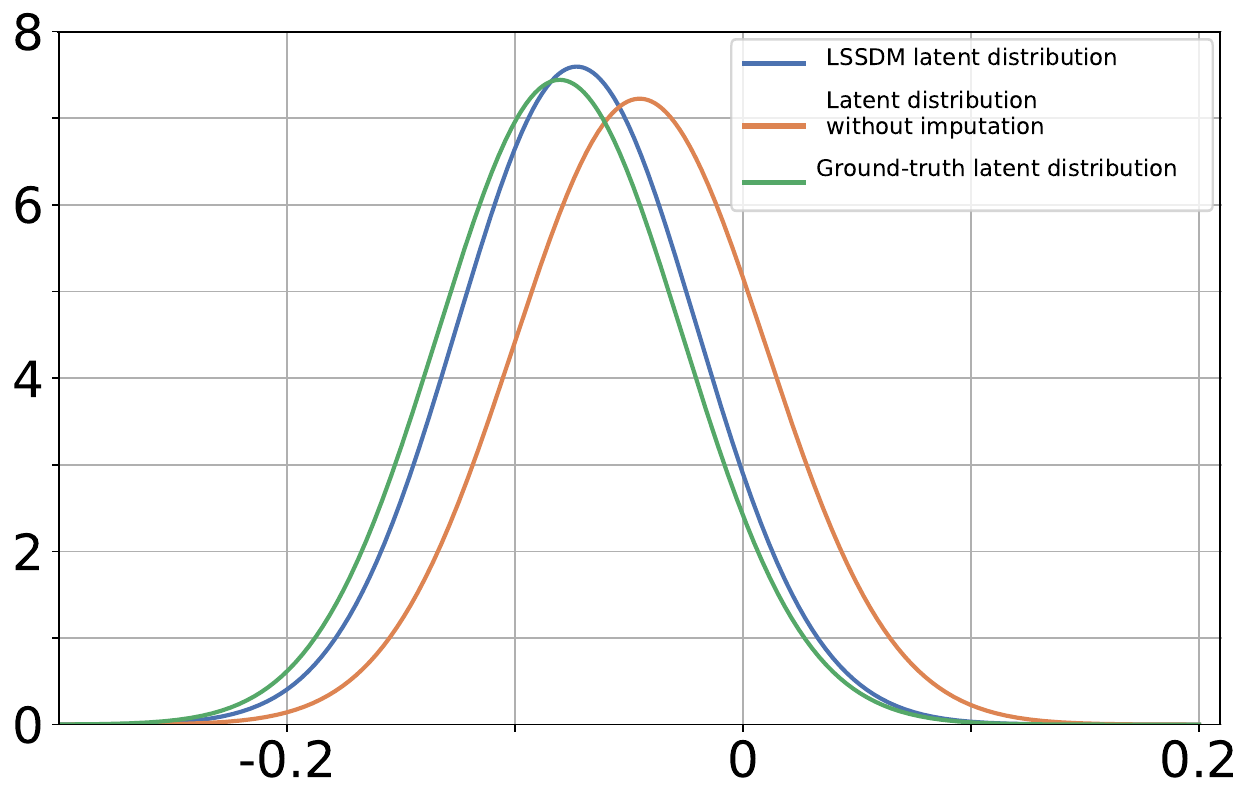} 
\caption{The latent distribution of different input on P12.}
\label{latent_state}
\end{figure}

To explore the generative mechanism, 50\% simulated missing and no simulated missing input on the P12 dataset into the trained VAE model and compared with the latent space distribution of LSSDM. The missing effect is tested, and we average all dimensions of $\mathbf{u}_\phi$ and $\mathbf{\Sigma}_\phi$ to plot the final latent Gaussian distributions, which are shown in Fig \ref{latent_state}. It demonstrates that the missing data can affect the latent space largely and cause the deviation of the latent space, which can not be neglected by the downstream task. However, LSSDM matches the no-simulated missing latent distribution better and provides insight into the latent distribution of the dataset, which is quite beneficial for downstream tasks.
\vspace{-1.0em}
\section{conclusion}
In this study, we propose LSSDM for MSTI. LSSDM learns the latent space of the dataset by GCN, and the coarse reconstructed missing values are obtained by Transformer and CNN. Then, the accurate imputation values can be achieved by inputting the reconstructed values into the diffusion model. By leveraging this innovative framework, we can obtain high-fidelity imputation with a better explanation of the imputation mechanism. Also, this model can utilize the unsupervised VGAE model to deal with the originally missing data without the label, which demonstrates more feasibility and practicality in a real application. Experimental results on real-world datasets indicate that LSSDM achieves superior imputation performance and uncertainty analysis.

\newpage



\begin{thebibliography}{00}

\bibitem{vaswani2017attention} A. Vaswani, ``Attention is all you need,'' Adv. Neural Inf. Process. Syst., Neural Information Processing Systems Foundation, vol. 5999, 2017.

\bibitem{cinifilling} A. Cini, I. Marisca, and C. Alippi, ``Filling the G\_ap\_s: Multivariate time series imputation by graph neural networks,'' in Proc. Int. Conf. Learn. Representations.

\bibitem{pakiyarajah2024irregularity} D. Pakiyarajah, E. Pavez, and A. Ortega, ``Irregularity-aware bandlimited approximation for graph signal interpolation,'' in Proc. IEEE Int. Conf. Acoust., Speech, Signal Process. (ICASSP), Apr. 2024, pp. 9801-9805.

\bibitem{liang2024physics} G. Liang, P. Tiwari, S. Nowaczyk, and S. Byttner, ``Higher-order spatio-temporal physics-incorporated graph neural network for multivariate time series imputation,'' in Proc. 33rd ACM Int. Conf. Inf. Knowl. Manag., 2024.

\bibitem{silva2012predicting} I. Silva, G. Moody, D. J. Scott, L. A. Celi, and R. G. Mark, ``Predicting in-hospital mortality of ICU patients: The physionet/computing in cardiology challenge 2012,'' in Proc. Comput. Cardiol., IEEE, 2012, pp. 245--248.

\bibitem{ho2020denoising} J. Ho, A. Jain, and P. Abbeel, ``Denoising diffusion probabilistic models,'' Adv. Neural Inf. Process. Syst., vol. 33, pp. 6840--6851, 2020.

\bibitem{fan2021multi} J. Fan, ``Multi-mode deep matrix and tensor factorization,'' in Proc. Int. Conf. Learn. Representations, 2021.

\bibitem{alcarazdiffusion} J. L. Alcaraz and N. Strodthoff, ``Diffusion-based time series imputation and forecasting with structured state space models,'' Trans. Mach. Learn. Res.

\bibitem{qiu2020multivariate} J. Qiu, S. R. Jammalamadaka, and N. Ning, ``Multivariate time series analysis from a Bayesian machine learning perspective,'' Ann. Math. Artif. Intell., vol. 88, pp. 1061--1082, 2020.

\bibitem{matheson1976scoring} J. E. Matheson and R. L. Winkler, ``Scoring rules for continuous probability distributions,'' Manage. Sci., vol. 22, no. 10, pp. 1087--1096, 1976.

\bibitem{xu2022pulseimpute} M. Xu, A. Moreno, S. Nagesh, V. Aydemir, D. Wetter, S. Kumar, and J. M. Rehg, ``Pulseimpute: A novel benchmark task for pulsative physiological signal imputation,'' Adv. Neural Inf. Process. Syst., vol. 35, pp. 26874--26888, 2022.

\bibitem{liu2023pristi} M. Liu, H. Huang, H. Feng, L. Sun, B. Du, and Y. Fu, ``Pristi: A conditional diffusion framework for spatiotemporal imputation,'' in Proc. IEEE 39th Int. Conf. Data Eng. (ICDE), 2023, pp. 1927--1939.

\bibitem{choi2023conditional} M. Choi and C. Lee, ``Conditional information bottleneck approach for time series imputation,'' in Proc. 12th Int. Conf. Learn. Representations, 2023.

\bibitem{abiri2019establishing} N. Abiri, B. Linse, P. Ed{\'e}n, and M. Ohlsson, ``Establishing strong imputation performance of a denoising autoencoder in a wide range of missing data problems,'' Neurocomputing, vol. 365, pp. 137--146, 2019.

\bibitem{viswarupan2024mixed} N. Viswarupan, G. Cheung, F. Lan, and M. S. Brown, ``Mixed graph signal analysis of joint image denoising/interpolation,'' in Proc. IEEE Int. Conf. Acoust., Speech, Signal Process. (ICASSP), Apr. 2024, pp. 9431-9435.


\bibitem{ding2024accurate} S. Ding, B. Xia, J. Sui, and D. Bu, ``Accurate interpolation of scattered data via learning relation graph,'' in Proc. IEEE Int. Conf. Acoust., Speech, Signal Process. (ICASSP), Apr. 2024, pp. 7290-7294.


\bibitem{nie2023imputeformer} T. Nie, G. Qin, W. Ma, Y. Mei, and J. Sun, ``ImputeFormer: Low rankness-induced transformers for generalizable spatiotemporal imputation,'' in Proc. 30th ACM SIGKDD Conf. Knowl. Discov. Data Min., 2024, pp. 2260--2271.

\bibitem{kipf2016semi} T. N. Kipf and M. Welling, ``Semi-supervised classification with graph convolutional networks,'' in Proc. 5th Int. Conf. Learn. Representations (ICLR), 2017. [Online]. Available: https://arxiv.org/abs/1609.02907

\bibitem{fortuin2020gp} V. Fortuin, D. Baranchuk, G. R{\"a}tsch, and S. Mandt, ``Gp-vae: Deep probabilistic time series imputation,'' in Proc. Int. Conf. Artif. Intell. Stat., PMLR, 2020, pp. 1651--1661.

\bibitem{cao2018brits} W. Cao, D. Wang, J. Li, H. Zhou, L. Li, and Y. Li, ``Brits: Bidirectional recurrent imputation for time series,'' Adv. Neural Inf. Process. Syst., vol. 31, 2018.

\bibitem{du2023saits} W. Du, D. C{\^o}t{\'e}, and Y. Liu, ``Saits: Self-attention-based imputation for time series,'' Expert Syst. Appl., vol. 219, p. 119619, 2023.

\bibitem{yi2016st} X. Yi, Y. Zheng, J. Zhang, and T. Li, ``ST-MVL: Filling missing values in geo-sensory time series data,'' in Proc. 25th Int. Joint Conf. Artif. Intell., 2016.

\bibitem{tashiro2021csdi} Y. Tashiro, J. Song, Y. Song, and S. Ermon, ``Csdi: Conditional score-based diffusion models for probabilistic time series imputation,'' Adv. Neural Inf. Process. Syst., vol. 34, pp. 24804--24816, 2021.

\bibitem{li2018dcrnn} Y. Li, R. Yu, C. Shahabi, and Y. Liu, ``Diffusion convolutional recurrent neural network: Data-driven traffic forecasting,'' in Proc. Int. Conf. Learn. Representations (ICLR), 2018. [Online]. Available: https://github.com/liyaguang/DCRNN

\bibitem{kong2020diffwave} Z. Kong, W. Ping, J. Huang, K. Zhao, and B. Catanzaro, ``DiffWave: A versatile diffusion model for audio synthesis,'' in Proc. Int. Conf. Learn. Representations (ICLR), 2021.

\end{thebibliography}
\end{document}